\documentclass[11pt,a4paper]{article}
\usepackage[hyperref]{naaclhlt2019}
\usepackage{times}
\usepackage{latexsym}
\usepackage{microtype}

\usepackage{graphicx}
\usepackage{arydshln}
\usepackage{caption}
\usepackage{subcaption}
\usepackage{amsmath}
\usepackage{amssymb}
\usepackage{multirow}
\usepackage{arydshln}
\usepackage{booktabs}
\usepackage{xcolor}
\usepackage{wrapfig}
\usepackage{hyperref}
\usepackage{url}

\aclfinalcopy 

\title{FAKTA: An Automatic End-to-End Fact Checking System}

\author{Moin Nadeem, Wei Fang, Brian Xu, Mitra Mohtarami, James Glass\\
  MIT Computer Science and Artificial Intelligence Laboratory \\ Cambridge, MA, USA \\
  {\tt \{mnadeem, weifang, bwxu, mitram, glass\}@mit.edu} \\}
\date{}

\begin{document}
\maketitle
\begin{abstract}
We present FAKTA which is a unified framework that integrates various components of a fact checking process: document retrieval from media sources with various types of reliability, stance detection of documents with respect to given claims, evidence extraction, and linguistic analysis. FAKTA predicts the factuality of given claims and provides evidence at the document and sentence level to explain its predictions.
\end{abstract}

\section{Introduction}
With the rapid increase of fake news in social media and its negative influence on people and public opinion~\cite{mihaylov-georgiev-nakov:2015:CoNLL, ACL2016:trolls,Vosoughi1146}, various organizations are now performing {\it manual} fact checking on suspicious claims. However, manual fact-checking is a time consuming and challenging process. As an alternative, researchers are investigating {\it automatic} fact checking which is a multi-step process and involves: (\emph{i})~retrieving potentially relevant documents for a given claim~\cite{AAAIFactChecking2018,R17-1046}, (\emph{ii})~checking the reliability of the media sources from which documents are retrieved
, (\emph{iii})~predicting the stance of each document with respect to the claim~\cite{mitra2018memory,brian:nips2018}, and finally (\emph{iv})~predicting factuality of given claims~\cite{AAAIFactChecking2018}. While previous works separately investigated individual components of the fact checking process, in this work, we present a unified framework titled FAKTA that integrates these components to not only predict the factuality of given claims, but also provide evidence at the document and sentence level to explain its predictions. To the best of our knowledge, FAKTA is the only system that offers such a capability.


\section{FAKTA}

\begin{figure*}[t]
\centering
\includegraphics[width=145mm,scale=2]{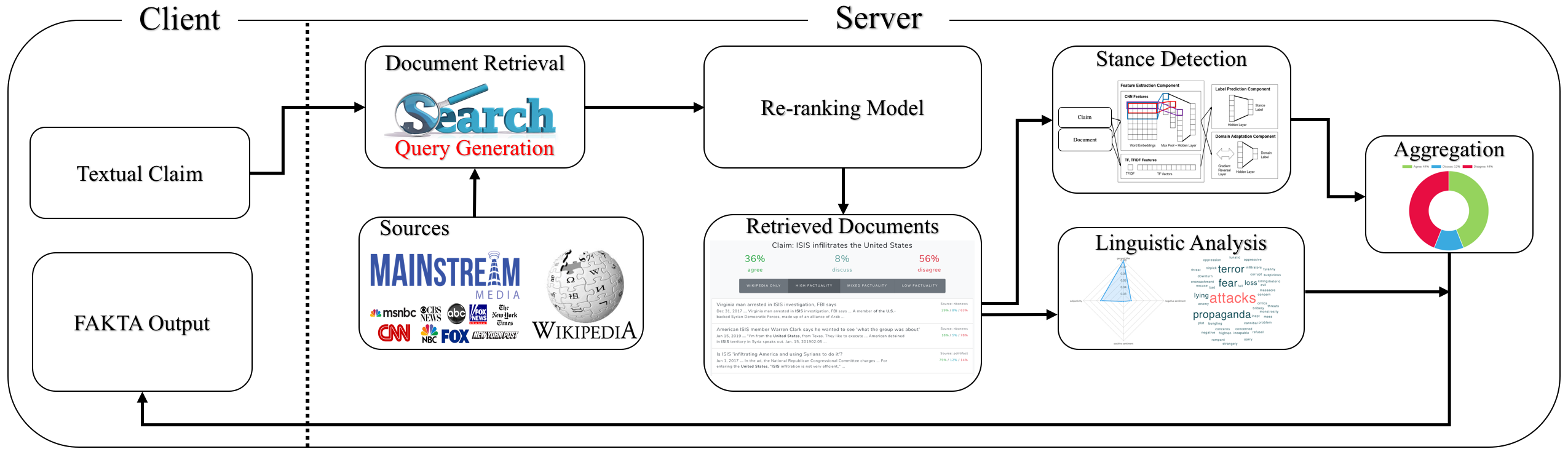}
\caption{The architecture of our FAKTA system.}
\label{fig:FAKTA-architecture}
\vspace{-7pt}
\end{figure*}

Figure~\ref{fig:FAKTA-architecture} illustrates the general architecture of FAKTA. The system is accessible via a Web browser and has two sides: client and server. When a user at the client side submits a textual claim for fact checking, the server handles the request by first passing it into the document retrieval component to retrieve a list of top-K relevant documents (see Section~\ref{Document-Retrieval}) from four types of sources: Wikipedia, highly-reliable, mixed reliability and low reliability mainstream media (see Section~\ref{Sources}). The retrieved documents are passed to the re-ranking model to refine the retrieval result (see Section~\ref{Document-Retrieval}). Then, the stance detection component detects the stance/perspective of each relevant document with respect to the claim, typically modeled using labels such as \textit{agree}, \textit{disagree} and \textit{discuss}. This component further provides rationales at the sentence level for explaining model predictions (see Section~\ref{Stance-Detection}). Each document is also passed to the linguistic analysis component to analyze the language of the document using different linguistic lexicons (see Section~\ref{Linguistic-Analysis}). Finally, the aggregation component combines the predictions of stance detection for all the relevant documents and makes a final decision about the factuality of the claim (see Section~\ref{Aggregation}). We describe the components below.

\subsection{Document Retrieval \& Re-ranking Model}\label{Document-Retrieval}
We first convert an input claim to a query by only considering its verbs, nouns and adjectives~\cite{PotthastHGTKRSS13}. 
Furthermore, claims often contain named entities (e.g., names of persons and organizations). We use the NLTK package to identify named entities in claims, and augment the initial query with all named entities from the claim's text. Ultimately, we generate queries of $5$--$10$ tokens, which we execute against a search engine. If the search engine does not retrieve any results for the query, we iteratively relax the query by dropping the final tokens one at a time.
We also use Apache Lucene\footnote{\url{https://lucene.apache.org}} to index and retrieve relevant documents from the $2017$ Wikipedia dump (see our experiments in Section~\ref{sec:results}). Furthermore, we use the Google API\footnote{\url{https://developers.google.com/custom-search}} to search across three pre-defined lists of media sources based on their factuality and reliability as explained in Section~\ref{Sources}. Finally, the re-ranking model of~\citeauthor{D18-1143}~(\citeyear{D18-1143}) is applied to select the top-K relevant documents.
This model uses all the POS tags in a claim that carry high discriminating power (NN, NNS, NNP, NNPS, JJ, CD) as keywords. The re-ranking model is defined as follows:
\begin{equation}
    f_{rank}=\frac{|match|}{|claim|} \times \frac{|match|}{|title|} \times score_{init},
\end{equation}
where $|claim|$, $|title|$, and $|match|$ are the counts of such POS tags in the claim, title of a document, both claim and title respectively, and $score_{init}$ is the initial ranking score computed by Lucene or ranking from Google API.

\subsection{Sources}\label{Sources}
While current search engines (e.g., Google, Bing, Yahoo) retrieve relevant documents for a given query from any media source, we retrieve relevant documents from four types of sources: Wikipedia, and high, mixed and low factual media. Journalists often spend considerable time verifying the reliability of their information sources~\cite{Popat:2017:TLE:3041021.3055133, NguyenKLW18}, and some fact-checking organizations have been producing lists of unreliable online news sources specified by their journalists. FAKTA utilizes information about news media listed on the Media Bias/Fact Check (MBFC) website\footnote{\url{https://mediabiasfactcheck.com}}, which contains manual annotations and analysis of the factuality of $2,500$ news websites. Our list from MBFC includes ~$1,300$ websites annotated by journalists as \textit{high} or \textit{very high}, ~$700$ websites annotated as \textit{low} and \textit{low-questionable}, and ~$500$ websites annotated as \textit{mixed} (i.e., containing both factually true and false information). Our document retrieval component retrieves documents from these three types of media sources (i.e., \textit{high}, \textit{mixed} and \textit{low}) along with Wikipedia that mostly contains factually-true information.

\begin{figure*}[t]
\centering
\includegraphics[width=141mm,scale=1]{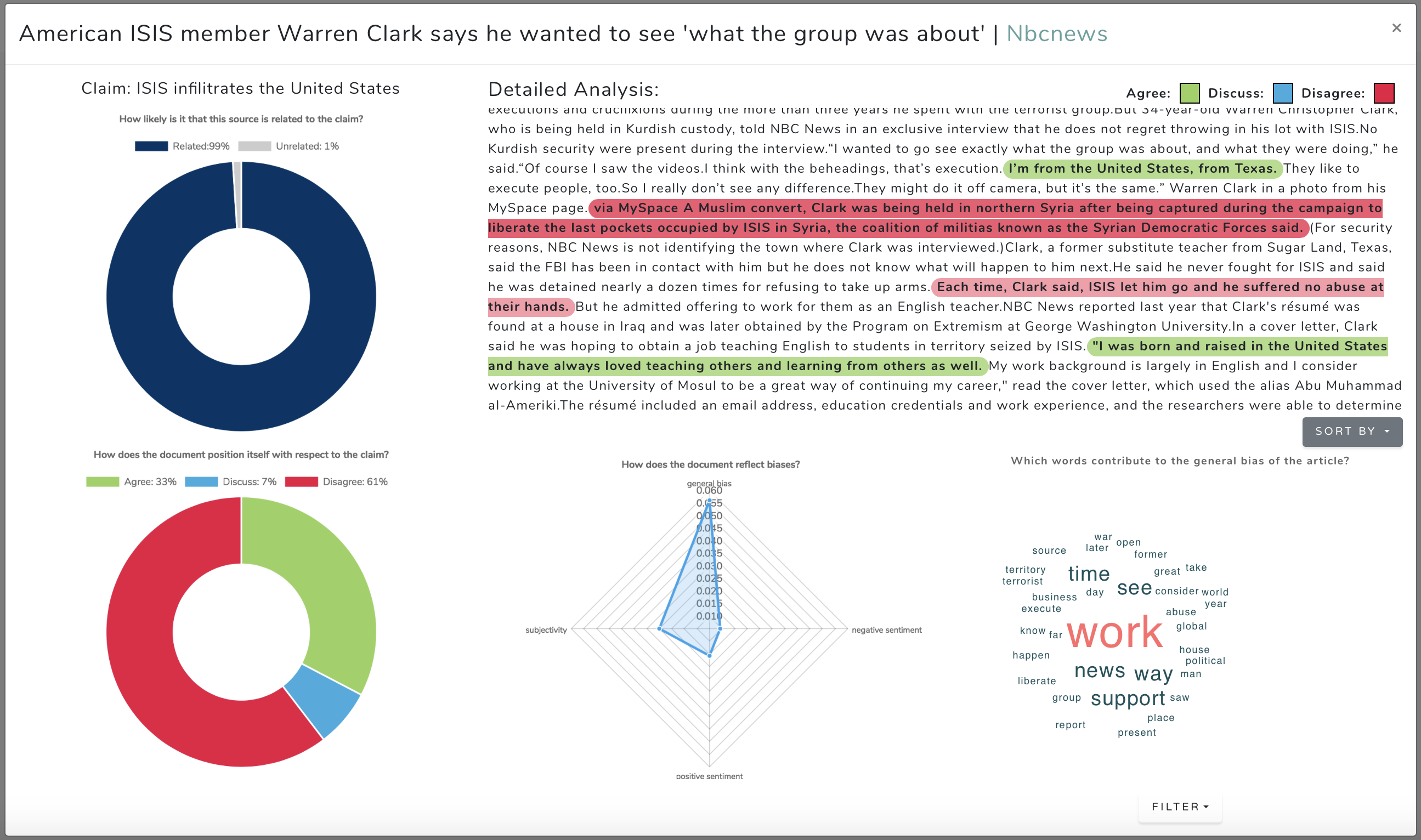}
\caption{Screenshot of FAKTA for a document retrieved for the claim ``ISIS infilitrates the United States."}
\label{fig:screenshot}
\vspace{-7pt}
\end{figure*}

\subsection{Stance Detection \& Evidence Extraction}\label{Stance-Detection}

In this work, we use our best model presented in~\cite{brian:nips2018} for stance detection. To the best of our knowledge, this model is the current state-of-the-art system on the Fake News Challenge (FNC) dataset.\footnote{\url{http://www.fakenewschallenge.org}} Our model combines Bag of Words (BOW) and Convolutional Neural Networks (CNNs) in a two-level \textit{hierarchy} scheme, where the first level predicts whether the label is \textit{related} or \textit{unrelated} (see Figure~\ref{fig:screenshot}, the top-left pie chart in FAKTA), and then related documents are passed to the second level to determine their stances, \textit{agree}, \textit{disagree}, and \textit{discuss} labels (see Figure~\ref{fig:screenshot}, the bottom-left pie chart in FAKTA). 
Our model is further supplemented with an adversarial domain adaptation technique which helps it overcome the limited size of labeled data when training through different domains.

To provide rationales for model prediction, FAKTA further processes each sentence in the document with respect to the claim and computes a stance score for each sentence. The relevant sentences in the document are then highlighted and color coded with respect to stance labels (see Figure~\ref{fig:screenshot}). FAKTA provides the option for re-ordering these rationales according to a specific stance label.

\subsection{Linguistic Analysis}\label{Linguistic-Analysis}
We analyze the language used in documents using the following linguistic markers: 

\noindent---\textit{Subjectivity lexicon}~\cite{Riloff:2003:LEP:1119355.1119369}: which contains weak and strong subjective terms (we only consider the strong subjectivity cues), 

\noindent---\textit{Sentiment cues}~\cite{Liu:2005:OOA:1060745.1060797}: which contains \textit{positive} and \textit{negative} sentiment cues, and 

\noindent---\textit{Wiki-bias lexicon}~\cite{Recasens:ACL:13}: which involves bias cues and controversial words (e.g., {\it abortion} and {\it execute}) extracted from the Neutral Point of View Wikipedia corpus~\cite{Recasens:ACL:13}.

Finally, we compute a score for the document using these cues according to Equation~\eqref{Ling-equation}, where for each lexicon type $L_i$ and document $D_j$, the frequency of the cues for $L_i$ in $D_j$ is normalized by the total number of words in $D_j$:
\begin{equation}\label{Ling-equation}\footnotesize
L_i(D_j) = \dfrac{\sum\limits_{cue \in L_i} {count(cue, D_j)}}{\sum\limits_{w_k \in D_j} {count(w_k, D_j)}}
\end{equation}

These scores are shown in a radar chart in Figure~\ref{fig:screenshot}. Furthermore, FAKTA provides the option to see a lexicon-specific word cloud of frequent words in each documents (see Figure~\ref{fig:screenshot}, the right side of the radar chart which shows the word cloud of Sentiment cues in the document).

\subsection{Aggregation}\label{Aggregation}
Stance Detection and Linguistic Analysis components are executed in parallel against all documents retrieved by our document retrieval component from each type of sources. All the stance scores are averaged across these documents, and the aggregated scores are shown for each {\it agree}, {\it disagree} and {\it discuss} categories at the top of the ranked list of retrieved documents. Higher agree score indicates the claim is factually true, and higher disagree score indicates the claim is factually false.


\section{Evaluation and Results}

We use the Fact Extraction and VERification (FEVER) dataset~\cite{FEVER} to evaluate our system. In FEVER, each claim is assigned to its relevant Wikipedia documents  with agree/disagree stances to the claim, and claims are labeled as \textit{supported} (SUP, i.e. factually true), \textit{refuted} (REF, i.e. factually false), and \textit{not enough information} (NEI, i.e., there is not any relevant document for the claim in Wikipedia). The data includes a total of ~$145$K claims, with around $80$K, $30$K and $35$K SUP, REF and NEI labels respectively.

\textit{Document Retrieval:}
Table~\ref{tab:Retrieval-results} shows results for document retrieval. We use various search and ranking algorithms that measure the similarity between each input claim as query and Web documents. Lines 1--11 in the table show the results when we use Lucene to index and search the data corpus with the following retrieval models: BM25~\cite{Robertson96okapiat} (Line 1), Classic based on the TF.IDF model (Line 2), and Divergence from Independence (DFI)~\cite{Kocabas:2014:NTW:2592838.2592866} (Line 3). We also use Divergence from Independence Randomness (DFR)~\cite{Amati:2002:PMI:582415.582416} with different term frequency normalization, such as the normalization provided by Dirichlet prior (DFR$_{H_3}$) (Line 4) or a Zipfian relation prior (DFR$_{z}$) (Line 5). We also consider Information Based (IB) models~\cite{Clinchant:2010:IMA:1835449.1835490} with
Log-logistic (IB$_{LL}$) (Line 6) or Smoothed power-law (IB$_{SPL}$) (Line 7) distributions. Finally, we consider LMDirichlet~\cite{Zhai:2001:SSM:383952.384019} (Line 8), and LMJelinek~\cite{Zhai:2001:SSM:383952.384019} with different settings for its hyperparameter (Lines 9--11). According to the resulting performance at different ranks \{1--20\}, we select the ranking algorithm DFR$_{z}$ (Lucene$_{DFR_{Z}}$) as our retrieval model.

In addition, Lines 12--13 show the results when claims are converted to queries as explained in Section~\ref{Document-Retrieval}. The results (Lines 5 and 12) show that Lucene performance decreases with query generation. This might be because the resulting queries become more abstract than their corresponding claims which may introduce some noise to the intended meaning of claims. 
However, Lines 14--15 show that our re-ranking model, explained in Section~\ref{Document-Retrieval}, can improve both Lucene and Google results.

\begin{table}[t]
\centering
\scalebox{0.70}{
\begin{tabular}{lcccc}
  \toprule
	\bf Model & \bf R@1 & \bf R@5 & \bf R@10 & \bf R@20\\
  \midrule
  1. \ \ \ \ \ \multirow{1}*
    {BM25} & 28.84 & 38.66 & 62.34 & 70.10 \\
  2. \ \ \ \ \ \multirow{1}*
    {Classic} & 9.14 & 23.10 & 31.65 & 40.70 \\
  3. \ \ \ \ \ \multirow{1}*
    {DFI} & 40.93 & 66.98 & 74.84 & 81.22 \\
  4. \ \ \ \ \ \multirow{1}*
    {$DFR_{\it \small H3}$} & 43.67 & 71.18 & 78.32 & 83.16 \\
  5. \ \ \ \ \ \multirow{1}* 
    {$DFR_{\it \small Z}$} & \underline{43.14} & \underline{71.17} & \underline{78.60} & \underline{83.88} \\
  6. \ \ \ \ \ \multirow{1}*
    {$IB_{\it \small LL}$} & 41.86 & 68.02 & 75.46 & 81.13 \\
  7. \ \ \ \ \ \multirow{1}*
    {$IB_{\it \small SPL}$} & 42.27 & 69.55 & 77.03 & 81.99 \\
  8. \ \ \ \ \ \multirow{1}*
    {LMDirichlet} & 39.00 & 68.86 & 77.39 & 83.04 \\
  9. \ \ \ \ \ \multirow{1}*
    {$LMJelinek_{\it \small 0.05}$} & 37.39 & 59.75 & 67.58 & 74.15 \\
  10. \ \ \ \multirow{1}*
    {$LMJelinek_{\it \small 0.10}$} & 37.30 & 59.85 & 67.58 & 74.44 \\
  11. \ \ \ \multirow{1}*
    {$LMJelinek_{\it \small 0.20}$} & 37.01 & 59.60 & 67.60 & 74.62 \\
  \midrule
  \multicolumn{5}{c}{\textbf{using Query Generation}} \\
  \hdashline[1.5pt/2pt]
    12. \ \ \ \multirow{1}*
      {$Lucene_{DFR_{\it \small Z}}$} & 40.70 & 68.48 & 76.21 & 81.93 \\
    13. \ \ \ \multirow{1}*
      {Google API} & 56.62 & 71.92 & 73.86 & 74.89 \\
  \midrule
    \multicolumn{5}{c}{\textbf{using Re-ranking Model}} \\
  \hdashline[1.5pt/2pt]
    14. \ \ \ \multirow{1}*
      {$Lucene_{DFR_{\it \small Z}}$} & \textbf{62.37} & \textbf{78.12} & \textbf{80.84} & \textbf{82.11} \\
  15. \ \ \ \multirow{1}*
    {Google API} & \underline{57.80} & \underline{72.10} & \underline{74.15} & \underline{74.89} \\
    \bottomrule
\end{tabular}} 
\caption{Results of document retrieval on FEVER.}
\label{tab:Retrieval-results}
\vspace{-10pt}
\end{table}

\textit{FAKTA Full Pipeline:}
The complete pipeline consists of document retrieval and re-ranking model (Section~\ref{Document-Retrieval}), stance detection and rationale extraction\footnote{We used Intel AI's Distiller~\cite{neta_zmora_2018_1297430} to compress the model.} (Section~\ref{Stance-Detection}) and aggregation model (Section~\ref{Aggregation}). Table~\ref{tab:FAKTA-results} shows the results for the full pipeline. Lines 1--3 show the results for all three SUP, REF, and NEI labels (3lbl) and Randomly Sampled (RS) documents from Wikipedia for the NEI label. 
We label claims as NEI if the most relevant document retrieved has a retrieval score less than a threshold, which was determined by tuning on development data.
Line 1 is the multi-layer perceptron (MLP) model presented in~\cite{DBLP:journals/corr/RiedelASR17}. Lines 2--3 are the results for our system when using Lucene (L) and Google API (G) for document retrieval. The results show that our system achieves the highest performance on both F$_{1(Macro)}$ and accuracy (Acc) using Google as retrieval engine. We repeat our experiments when considering only SUP and REF labels (2lbl) and the results are significantly higher than the results with 3lbl (Lines 4--5). 

\begin{table}[t]
\centering
\scalebox{0.70}{
\begin{tabular}{lcccc}
  \toprule
	\bf Model & \bf Settings & \bf F$_{1(SUP/REF/NEI)}$ & \bf F$_{1(Macro)}$ & \bf Acc.\\
  \midrule
  1. \ \ \ \multirow{1}*
    {MLP} & 3lbl/RS & - & - & 40.63\\
  2. \ \ \ \multirow{1}*
    {FAKTA} & L/3lbl/RS & 41.33/23.55/44.79 & 36.56 & 38.76 \\
  3. \ \ \ \multirow{1}*
    {FAKTA} & G/3lbl/RS & \underline{47.49/43.01/28.17} & \underline{39.65} & \underline{41.21} \\
  \hdashline[1.5pt/2pt]
  4. \ \ \ \multirow{1}*
    {FAKTA} & L/2lbl & 58.33/57.71/- & 58.02 & 58.03 \\
  5. \ \ \ \multirow{1}*
    {FAKTA} & G/2lbl & \underline{58.96/59.74/-} & \underline{59.35} & \underline{59.35} \\
\bottomrule
\end{tabular}}
\caption{FAKTA full pipeline Results on FEVER.}
\label{tab:FAKTA-results}
\vspace{-10pt}
\end{table}

\label{sec:results}

\section{The System in Action}
The current version of FAKTA\footnote{\url{http://fakta.mit.edu}} and its short introduction video\footnote{\url{http://fakta.mit.edu/video}} and source code\footnote{\url{https://github.com/moinnadeem/fakta}} are available online. FAKTA consists of three views:

\noindent---\textit{The text entry view}: to enter a claim to be checked for factuality.

\noindent---\textit{Overall result view}: includes four lists of retrieved documents from four factuality types of sources: Wikipedia, and high-, mixed-, and low-factuality media (Section~\ref{Sources}). For each list, the final factuality score for the input claim is shown at the top of the page (Section~\ref{Aggregation}), and the stance detection score for each document appears beside it.

\noindent---\textit{Document result view}: when selecting a retrieved document, FAKTA shows the text of the document and highlights its important sentences according to their stance scores with respect to the claim. The stance detection results for the document are further shown as pie chart at the left side of the view (Section~\ref{Stance-Detection}), and the linguistic analysis is shown at the bottom of the view (Section~\ref{Linguistic-Analysis}).

\section{Related Work}

Automatic fact checking~\cite{brian:nips2018} centers on
evidence extraction for given claims,
reliability evaluation of media sources~\cite{D18-1389},
stance detection of documents with respect to claims~\cite{mitra2018memory,brian:nips2018,baly2018integrating}, and
fact checking of claims~\cite{AAAIFactChecking2018}.
These steps correspond to different Natural Language Processing (NLP) and Information Retrieval (IR) tasks
including information extraction and question answering~\cite{shiralkar2017finding}. 
Veracity inference has been mostly approached as 
text classification problem and mainly tackled by developing linguistic, stylistic, and semantic features~\cite{rashkin2017truth,AAAIFactChecking2018,nakov2017do}, as well as using information from \emph{external} sources~\cite{AAAIFactChecking2018,R17-1046}.

These steps are typically handled in isolation. For example, previous works~\cite{liar-P17-2067,Nicole:nips2018} proposed algorithms to predict factuality of claims by mainly focusing on only input claims (i.e., step (\emph{iv}) and their metadata information (e.g., the speaker of the claim). In addition, recent works on the Fact Extraction and VERification (FEVER)~\cite{FEVER} has focused on a specific domain (e.g., Wikipedia). 

To the best of our knowledge, there is currently no end-to-end systems for fact checking which 
can search through Wikipedia and mainstream media sources across the Web to fact check given claims. To address these gaps, our FAKTA system covers all fact-checking steps and can search across different sources, predict the factuality of claims, and present a set of evidence to explain its prediction. 

\section{Conclusion}
We have presented FAKTA--an online system for automatic end-to-end fact checking of claims. FAKTA can assist individuals and professional fact-checkers to check the factuality of claims by presenting relevant documents and rationales as evidence for its predictions. In future work, we plan to improve FAKTA's underlying components (e.g., stance detection), extend FAKTA to cross-lingual settings, and incorporate temporal information for fact checking.

\section*{Acknowledgments}
We thank anonymous reviewers for their insightful comments, suggestions, and feedback. This research was supported in part by HBKU Qatar Computing Research Institute (QCRI), DSTA of Singapore, and Intel AI.

\bibliography{naaclhlt2019}

\begin{thebibliography}{29}
\expandafter\ifx\csname natexlab\endcsname\relax\def\natexlab#1{#1}\fi

\bibitem[{Amati and Van~Rijsbergen(2002)}]{Amati:2002:PMI:582415.582416}
Gianni Amati and Cornelis~Joost Van~Rijsbergen. 2002.
\newblock Probabilistic models of information retrieval based on measuring the
  divergence from randomness.
\newblock \emph{ACM Transactions on Information Systems (TOIS)},
  20(4):357--389.

\bibitem[{Baly et~al.(2018{\natexlab{a}})Baly, Karadzhov, Alexandrov, Glass,
  and Nakov}]{D18-1389}
Ramy Baly, Georgi Karadzhov, Dimitar Alexandrov, James Glass, and Preslav
  Nakov. 2018{\natexlab{a}}.
\newblock Predicting factuality of reporting and bias of news media sources.
\newblock In \emph{Proceedings of the 2018 Conference on Empirical Methods in
  Natural Language Processing}, pages 3528--3539. Association for Computational
  Linguistics.

\bibitem[{Baly et~al.(2018{\natexlab{b}})Baly, Mohtarami, Glass, M\`arquez,
  Moschitti, and Nakov}]{baly2018integrating}
Ramy Baly, Mitra Mohtarami, James Glass, Llu\'is M\`arquez, Alessandro
  Moschitti, and Preslav Nakov. 2018{\natexlab{b}}.
\newblock Integrating stance detection and fact checking in a unified corpus.
\newblock In \emph{Proceedings of the 16th Annualw Conference of the North
  American Chapter of the Association for Computational Linguistics},
  NAACL-HLT~'18, New Orleans, LA, USA.

\bibitem[{Clinchant and Gaussier(2010)}]{Clinchant:2010:IMA:1835449.1835490}
St{\'e}phane Clinchant and Eric Gaussier. 2010.
\newblock Information-based models for ad hoc ir.
\newblock In \emph{Proceedings of the 33rd International ACM SIGIR Conference
  on Research and Development in Information Retrieval}, SIGIR'10, pages
  234--241, New York, NY, USA. ACM.

\bibitem[{Karadzhov et~al.(2017)Karadzhov, Nakov, M{\`a}rquez,
  Barr{\'o}n-Cede{\~{n}}o, and Koychev}]{R17-1046}
Georgi Karadzhov, Preslav Nakov, Llu{\'i}s M{\`a}rquez, Alberto
  Barr{\'o}n-Cede{\~{n}}o, and Ivan Koychev. 2017.
\newblock Fully automated fact checking using external sources.
\newblock In \emph{Proceedings of the International Conference Recent Advances
  in Natural Language Processing, RANLP 2017}, pages 344--353. INCOMA Ltd.

\bibitem[{Kocaba\c{s} et~al.(2014)Kocaba\c{s}, Din\c{c}er, and
  Karao\u{g}lan}]{Kocabas:2014:NTW:2592838.2592866}
\.{I}lker Kocaba\c{s}, Bekir~Taner Din\c{c}er, and Bahar Karao\u{g}lan. 2014.
\newblock A nonparametric term weighting method for information retrieval based
  on measuring the divergence from independence.
\newblock \emph{Inf. Retr.}, 17(2):153--176.

\bibitem[{Lee et~al.(2018)Lee, Wu, and Fung}]{D18-1143}
Nayeon Lee, Chien-Sheng Wu, and Pascale Fung. 2018.
\newblock Improving large-scale fact-checking using decomposable attention
  models and lexical tagging.
\newblock In \emph{Proceedings of the 2018 Conference on Empirical Methods in
  Natural Language Processing}, pages 1133--1138. Association for Computational
  Linguistics.

\bibitem[{Liu et~al.(2005)Liu, Hu, and Cheng}]{Liu:2005:OOA:1060745.1060797}
Bing Liu, Minqing Hu, and Junsheng Cheng. 2005.
\newblock Opinion observer: Analyzing and comparing opinions on the web.
\newblock In \emph{Proceedings of the 14th International Conference on World
  Wide Web}, pages 342--351, Chiba, Japan.

\bibitem[{Mihaylov et~al.(2015)Mihaylov, Georgiev, and
  Nakov}]{mihaylov-georgiev-nakov:2015:CoNLL}
Todor Mihaylov, Georgi Georgiev, and Preslav Nakov. 2015.
\newblock Finding opinion manipulation trolls in news community forums.
\newblock In \emph{Proceedings of the Nineteenth Conference on Computational
  Natural Language Learning}, pages 310--314. Association for Computational
  Linguistics.

\bibitem[{Mihaylov and Nakov(2016)}]{ACL2016:trolls}
Todor Mihaylov and Preslav Nakov. 2016.
\newblock Hunting for troll comments in news community forums.
\newblock In \emph{Proceedings of the 54th Annual Meeting of the Association
  for Computational Linguistics}, pages 399--405, Berlin, Germany.

\bibitem[{Mihaylova et~al.(2018)Mihaylova, Nakov, Marquez, Barron-Cedeno,
  Mohtarami, Karadzhov, and Glass}]{AAAIFactChecking2018}
Tsvetomila Mihaylova, Preslav Nakov, Lluis Marquez, Alberto Barron-Cedeno,
  Mitra Mohtarami, Georgi Karadzhov, and James Glass. 2018.
\newblock Fact checking in community forums.
\newblock In \emph{Proceedings of the Thirty-Second {AAAI} Conference on
  Artificial Intelligence}, pages 5309--5316, New Orleans, LA, USA.

\bibitem[{Mohtarami et~al.(2018)Mohtarami, Baly, Glass, Nakov, M\`arquez, and
  Moschitti}]{mitra2018memory}
Mitra Mohtarami, Ramy Baly, James Glass, Preslav Nakov, Llu\'is M\`arquez, and
  Alessandro Moschitti. 2018.
\newblock Automatic stance detection using end-to-end memory networks.
\newblock In \emph{Proceedings of the 16th Annualw Conference of the North
  American Chapter of the Association for Computational Linguistics},
  NAACL-HLT~'18, New Orleans, LA, USA.

\bibitem[{Nakov et~al.(2017)Nakov, Mihaylova, Marquez, Shiroya, and
  Koychev}]{nakov2017do}
Preslav Nakov, Tsvetomila Mihaylova, Llu{\i}s Marquez, Yashkumar Shiroya, and
  Ivan Koychev. 2017.
\newblock Do not trust the trolls: Predicting credibility in community question
  answering forums.
\newblock In \emph{Proceedings of the International Conference Recent Advances
  in Natural Language Processing (RANLP)}, pages 551--560.

\bibitem[{Nguyen et~al.(2018)Nguyen, Kharosekar, Lease, and
  Wallace}]{NguyenKLW18}
An~Nguyen, Aditya Kharosekar, Matthew Lease, and Byron Wallace. 2018.
\newblock An interpretable joint graphical model for fact-checking from crowds.
\newblock In \emph{AAAI Conference on Artificial Intelligence}.

\bibitem[{O’Brien et~al.(2018)O’Brien, Latessa, Evangelopoulos, and
  Boix}]{Nicole:nips2018}
Nicole O’Brien, Sophia Latessa, Georgios Evangelopoulos, and Xavier Boix.
  2018.
\newblock The language of fake news: Opening the black-box of deep learning
  based detectors.
\newblock In \emph{Proceedings of the Thirty-second Annual Conference on Neural
  Information Processing Systems (NeurIPS)--AI for Social Good}.

\bibitem[{Popat et~al.(2017)Popat, Mukherjee, Str\"{o}tgen, and
  Weikum}]{Popat:2017:TLE:3041021.3055133}
Kashyap Popat, Subhabrata Mukherjee, Jannik Str\"{o}tgen, and Gerhard Weikum.
  2017.
\newblock Where the truth lies: Explaining the credibility of emerging claims
  on the web and social media.
\newblock In \emph{Proceedings of the 26th International Conference on World
  Wide Web Companion}, WWW '17 Companion, pages 1003--1012, Republic and Canton
  of Geneva, Switzerland.

\bibitem[{Potthast et~al.(2013)Potthast, Hagen, Gollub, Tippmann, Kiesel,
  Rosso, Stamatatos, and Stein}]{PotthastHGTKRSS13}
Martin Potthast, Matthias Hagen, Tim Gollub, Martin Tippmann, Johannes Kiesel,
  Paolo Rosso, Efstathios Stamatatos, and Benno Stein. 2013.
\newblock Overview of the 5th international competition on plagiarism
  detection.
\newblock In \emph{Working Notes for {CLEF} 2013 Conference , Valencia, Spain,
  September 23-26, 2013.}

\bibitem[{Rashkin et~al.(2017)Rashkin, Choi, Jang, Volkova, and
  Choi}]{rashkin2017truth}
Hannah Rashkin, Eunsol Choi, Jin~Yea Jang, Svitlana Volkova, and Yejin Choi.
  2017.
\newblock Truth of varying shades: Analyzing language in fake news and
  political fact-checking.
\newblock In \emph{Proceedings of the 2017 Conference on Empirical Methods in
  Natural Language Processing}, pages 2921--2927.

\bibitem[{Recasens et~al.(2013)Recasens, Danescu-Niculescu-Mizil, and
  Jurafsky}]{Recasens:ACL:13}
Marta Recasens, Cristian Danescu-Niculescu-Mizil, and Dan Jurafsky. 2013.
\newblock Linguistic models for analyzing and detecting biased language.
\newblock In \emph{Proceedings of the 51st Annual Meeting of the Association
  for Computational Linguistics}, pages 1650--1659, Sofia, Bulgaria.

\bibitem[{Riedel et~al.(2017)Riedel, Augenstein, Spithourakis, and
  Riedel}]{DBLP:journals/corr/RiedelASR17}
Benjamin Riedel, Isabelle Augenstein, Georgios~P. Spithourakis, and Sebastian
  Riedel. 2017.
\newblock A simple but tough-to-beat baseline for the fake news challenge
  stance detection task.
\newblock \emph{CoRR}, abs/1707.03264.

\bibitem[{Riloff and Wiebe(2003)}]{Riloff:2003:LEP:1119355.1119369}
Ellen Riloff and Janyce Wiebe. 2003.
\newblock Learning extraction patterns for subjective expressions.
\newblock In \emph{Proceedings of the Conference on Empirical Methods in
  Natural Language Processing}, pages 105--112, Sapporo, Japan.

\bibitem[{Robertson et~al.(1994)Robertson, Walker, Jones, Hancock-Beaulieu, and
  Gatford}]{Robertson96okapiat}
Stephen~E. Robertson, Steve Walker, Susan Jones, Micheline Hancock-Beaulieu,
  and Mike Gatford. 1994.
\newblock Okapi at trec-3.
\newblock In \emph{TREC}, pages 109--126. National Institute of Standards and
  Technology (NIST).

\bibitem[{Shiralkar et~al.(2017)Shiralkar, Flammini, Menczer, and
  Ciampaglia}]{shiralkar2017finding}
Prashant Shiralkar, Alessandro Flammini, Filippo Menczer, and Giovanni~Luca
  Ciampaglia. 2017.
\newblock Finding streams in knowledge graphs to support fact checking.
\newblock \emph{arXiv preprint arXiv:1708.07239}.

\bibitem[{Thorne et~al.(2018)Thorne, Vlachos, Christodoulopoulos, and
  Mittal}]{FEVER}
James Thorne, Andreas Vlachos, Christos Christodoulopoulos, and Arpit Mittal.
  2018.
\newblock Fever: a large-scale dataset for fact extraction and verification.
\newblock In \emph{Proceedings of the 2018 Conference of the North American
  Chapter of the Association for Computational Linguistics (HLT-NAACL)}, pages
  809--819.

\bibitem[{Vosoughi et~al.(2018)Vosoughi, Roy, and Aral}]{Vosoughi1146}
Soroush Vosoughi, Deb Roy, and Sinan Aral. 2018.
\newblock The spread of true and false news online.
\newblock \emph{Science}, 359(6380):1146--1151.

\bibitem[{Wang(2017)}]{liar-P17-2067}
William~Yang Wang. 2017.
\newblock ``liar, liar pants on fire'': A new benchmark dataset for fake news
  detection.
\newblock In \emph{Proceedings of the 55th Annual Meeting of the Association
  for Computational Linguistics (Volume 2: Short Papers)}, pages 422--426.
  Association for Computational Linguistics.

\bibitem[{Xu et~al.(2018)Xu, Mohtarami, and Glass}]{brian:nips2018}
Brian Xu, Mitra Mohtarami, and James Glass. 2018.
\newblock Adversarial doman adaptation for stance detection.
\newblock In \emph{Proceedings of the Thirty-second Annual Conference on Neural
  Information Processing Systems (NeurIPS)--Continual Learning}.

\bibitem[{Zhai and Lafferty(2001)}]{Zhai:2001:SSM:383952.384019}
Chengxiang Zhai and John Lafferty. 2001.
\newblock A study of smoothing methods for language models applied to ad hoc
  information retrieval.
\newblock In \emph{Proceedings of the 24th Annual International ACM SIGIR
  Conference on Research and Development in Information Retrieval}, SIGIR'01,
  pages 334--342, New York, NY, USA. ACM.

\bibitem[{Zmora et~al.(2018)Zmora, Jacob, and Novik}]{neta_zmora_2018_1297430}
Neta Zmora, Guy Jacob, and Gal Novik. 2018.
\newblock Neural network distiller.
\newblock Available at https://doi.org/10.5281/zenodo.1297430.

\end{thebibliography}
\bibliographystyle{acl_natbib}


\end{document}